\documentclass{article}

\usepackage{microtype}
\usepackage{graphicx}
\usepackage{subfigure}
\usepackage{booktabs} 
\usepackage{paralist} 

\usepackage{hyperref}
\usepackage[dvipsnames]{xcolor}

\newcommand{\care}[1]{{\color{black}#1}}

\definecolor{darkred}{RGB}{200, 90, 20}

\usepackage[accepted]{icml2024}

\PassOptionsToPackage{sort}{natbib}

\usepackage{tcolorbox}
\tcbuselibrary{theorems}
\tcbuselibrary{magazine}

\usepackage{amsmath}
\usepackage{amssymb}
\usepackage{mathtools}
\usepackage{amsthm}

\usepackage[capitalize,noabbrev]{cleveref}

\theoremstyle{plain}

\theoremstyle{definition}

\theoremstyle{remark}

\theoremstyle{plain}
\newtheorem{problem}{Open problem}
\newtheorem{direction}{Research direction}

\usepackage[textsize=tiny]{todonotes}

\usepackage{newunicodechar}

\DeclareRobustCommand{\okina}{%
  \raisebox{\dimexpr\fontcharht\font`A-\height}{%
    \scalebox{0.8}{`}%
  }%
}
\newunicodechar{ʻ}{\okina}

\icmltitlerunning{\care{Position: Topological Deep Learning is the New Frontier for Relational Learning}}

\begin{document}

\twocolumn[
\icmltitle{\care{Position: Topological Deep Learning is the New Frontier for Relational Learning}}




\begin{icmlauthorlist}
\icmlauthor{Theodore Papamarkou}{uom}
\icmlauthor{Tolga Birdal}{uoi}
\icmlauthor{Michael Bronstein}{uoo}
\icmlauthor{Gunnar Carlsson}{uost,bluelightai}
\icmlauthor{Justin Curry}{uoa}
\icmlauthor{Yue Gao}{uot}
\icmlauthor{Mustafa Hajij}{usf}
\icmlauthor{Roland Kwitt}{uosa}
\icmlauthor{Pietro Li\`{o}}{uoc}
\icmlauthor{Paolo Di Lorenzo}{uos}
\icmlauthor{Vasileios Maroulas}{utk}
\icmlauthor{Nina Miolane}{ucsb}
\icmlauthor{Farzana Nasrin}{uoh}
\icmlauthor{Karthikeyan Natesan Ramamurthy}{ibm}
\icmlauthor{Bastian Rieck}{hm,tum}
\icmlauthor{Simone Scardapane}{uos}
\icmlauthor{Michael T. Schaub}{aachen}
\icmlauthor{Petar Veli\v{c}kovi\'{c}}{deepmind,uoc}
\icmlauthor{Bei Wang}{uou}
\icmlauthor{Yusu Wang}{ucsd}
\icmlauthor{Guo-Wei Wei}{msu}
\icmlauthor{Ghada Zamzmi}{usfl}
\end{icmlauthorlist}

\icmlaffiliation{uom}{Department of Mathematics, The University of Manchester, Manchester, UK.}
\icmlaffiliation{uoi}{Department of Computing, Imperial College London, London, UK.}
\icmlaffiliation{uoo}{Department of Computer Science, University of Oxford, Oxford, UK.}
\icmlaffiliation{uost}{Department of Mathematics, Stanford University, Stanford, USA.}
\icmlaffiliation{bluelightai}{BlueLightAI Inc, USA.}
\icmlaffiliation{uoa}{University at Albany, New York, USA.}
\icmlaffiliation{uot}{School of Software, Tsinghua University, Beijing, China.}
\icmlaffiliation{usf}{University of San Francisco, San Francisco, USA.}
\icmlaffiliation{uosa}{Department of Artificial Intelligence and Human Interfaces,
University of Salzburg, Austria.}
\icmlaffiliation{uoc}{Department of Computer Science and Technology,
University of Cambridge, Cambridge, UK.}
\icmlaffiliation{uos}{Department of Information Engineering, Electronics and Telecommunications,
Sapienza University of Rome, Rome, Italy.}
\icmlaffiliation{utk}{Department of Mathematics, University of Tennessee, Knoxville, USA.}
\icmlaffiliation{ucsb}{Department of Electrical and Computer Engineering, UC Santa Barbara, Santa Barbara, USA.}
\icmlaffiliation{uoh}{Department of Mathematics,
University of Hawaiʻi at M\={a}noa, Hawaiʻi, USA.}
\icmlaffiliation{ibm}{IBM Corporation New York, USA.}
\icmlaffiliation{hm}{Helmholtz Munich, Munich Germany.}
\icmlaffiliation{tum}{Technical University of Munich, Munich Germany.}
\icmlaffiliation{aachen}{RWTH Aachen University, Aachen, Germany.}
\icmlaffiliation{deepmind}{Google DeepMind.}
\icmlaffiliation{uou}{School of Computing, University of Utah, Utah, USA.}
\icmlaffiliation{ucsd}{Computer Science and Engineering Department, University of California San Diego, San Diego, USA.}
\icmlaffiliation{msu}{Department  of  Mathematics, Michigan State University, East Lansing, Michigan, USA.}
\icmlaffiliation{usfl}{University of South Florida, Florida, USA}

\icmlcorrespondingauthor{Theodore Papamarkou}{theo.papamarkou@manchester.ac.uk}

\icmlkeywords{Geometric deep learning, graph representation learning, topological deep learning, topological neural networks}

\vskip 0.3in
]



\printAffiliationsAndNotice{}  

\begin{figure*}
\centering
\includegraphics[width=1\textwidth]{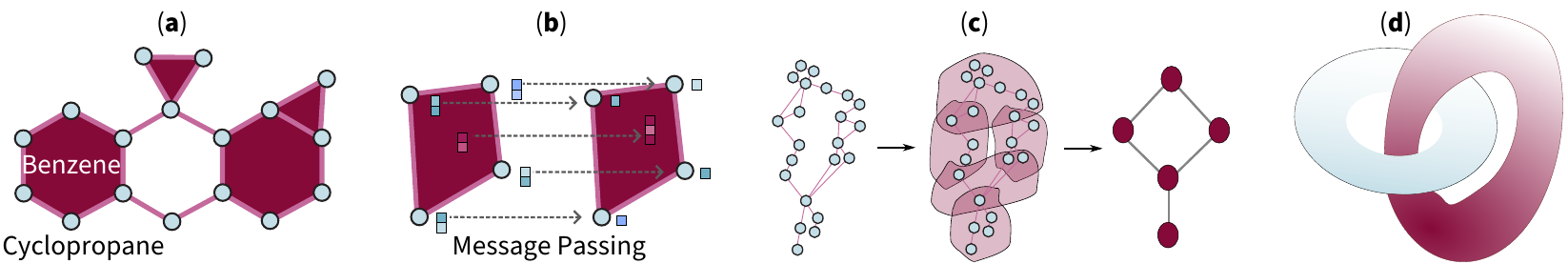}
\vspace{-5mm}
\caption{Topological spaces enrich deep learning methods from a variety of perspectives. (a): A topological space, modeled as a cell complex, enables a flexible molecular representation.
\care{This} representation can \care{improve} the performance of a deep learning model supported on this space~\citep{bodnar2021weisfeiler}. (b): \care{Topological neural networks enable the processing of data, for example via higher-order message-passing schemes on a topological space. These networks find a wide range of applications, from computer graphics to drug discovery~\citep{hajij2023tdl}}. (c): Topological spaces allow hierarchical representations of the underlying data that naturally correspond to pooling operations in deep learning~\citep{hajij2023tdl}. (d): \care{The} topological characteristics of the underlying data are crucial when selecting a neural network architecture. Data from a knotted structure in $\mathbb{R}^3$, such as the one shown, cannot be embedded in $\mathbb{R}^2$ with a single layer multilayer perceptron (MLP) from $\mathbb{R}^3$ to $\mathbb{R}^2$ \cite{olah2014neural}.\vspace{-4mm}}
\label{fig:img1}
\end{figure*}

\begin{abstract}
Topological deep learning (TDL) is a rapidly evolving field that uses topological features to understand and design deep learning models. This paper posits that \care{TDL is the new frontier for relational learning. TDL} may complement graph representation learning and geometric deep learning by incorporating topological concepts, and can thus provide a natural choice for various machine learning settings. To this end, this paper discusses open problems in TDL, ranging from practical benefits to theoretical foundations. For each problem, it outlines potential solutions and future research opportunities. At the same time, this paper serves as an invitation to the scientific community to actively participate in TDL research to unlock the potential of this emerging field.
\end{abstract}

\section{Introduction}

Traditional machine learning often assumes that the observed data of interest are supported on a linear vector space and can be described by a set of feature vectors. However, there is growing awareness that, in many cases, this viewpoint is insufficient to describe several data within the real world. For example, molecules may be described more appropriately by graphs than feature vectors. Other examples include three-dimensional objects represented by meshes, as encountered in computer graphics and geometry processing, or data supported on top of a complex social network of interrelated actors. Hence, there has been an increased interest in importing concepts from geometry and topology into the usual machine learning pipelines to gain further insights into such types of data in a systematic way. 

One of the most prominent examples of this research direction is geometric deep learning (GDL); see~\citep{bronstein2021geometric,zhou2020graph,wu2020comprehensive,nguyen2019dg}. GDL aims to generalize neural networks to non-Euclidean domains, including manifolds and graphs. Graph neural networks (GNNs) form a central pillar within GDL~\citep{wu2020comprehensive, zhou2020graph}.

Topology is concerned with the study of properties that remain invariant under continuous deformations, and affords a powerful lens through which the global structure of data can be discerned.
\care{By characterizing topological features (including connected components, loops, and voids across multiple scales), topological tools such as persistent homology~\citep{Carlsson2009,EdelsbrunnerHarer2010} have become powerful methods to capture essential structures and patterns that elude conventional methods.}
The ability to quantify the topological signatures of data can enhance the robustness of traditional machine learning models, enabling them to discern a greater variety of meaningful structures in diverse and complex datasets. This use of topological ideas, which often runs under the umbrella of topological data analysis (TDA), is by now a well-established field of research. For example, persistent homology has led to victories in the D3R Grand Challenges, a worldwide annual competition series in computer-aided drug design~ \citep{nguyen2019mathematical, nguyen2020mathdl}.
\care{Moreover, topological descriptors such as contour trees, merge trees, Reeb graphs, mapper, Morse and Morse-Smale complexes are established tools in scientific data analysis and visualization~\citep{BMWL20,Hu21a}.}

Various topological structures have been used as models to describe data in terms of its higher order relations, including cell complexes~\citep{hajijcell}, simplicial complexes~\citep{schaub2022signal}, sheaves~\citep{hansen2019toward}, and combinatorial complexes~\citep{hajij2023tdl}. Machine learning models have recently been designed to learn from data that are supported on these topological domains~\citep{billings2019simplex2vec,bunch2020simplicial,hajijcell,schaub2020random,roddenberry2021principled,roddenberry2021signal, schaub2021signal,giusti2023cell,yang2023convolutional}.

Beyond using topological features to describe data, it is also possible to leverage topological features to understand and control computational models. In other words, one may use topological methods to understand the flow of information within computational models, such as neural networks, to gain insight about the models' functional behavior. Furthermore, one may design deep learning architectures according to certain topological notions to enhance or constrain the type of computations that can be performed, for example, by enforcing message-passing schemes that are compliant with certain topological invariants or promoting desirable topological properties of learned representations.

Although there is an overlap between the three ways described above to incorporate topological notions into machine learning, this paper focuses primarily on the latter two points of view, which hold great potential for machine learning applications and are not yet well established, in contrast to `traditional' TDA~\citep{EdelsbrunnerHarer2010,Carlsson2009,DW22,ghrist2014elementary}. The discussion is particularly concerned with deep learning models, due to their broad real-world adoption. So, the scope of this paper is dedicated to the nascent field of topological deep learning (TDL), highlighting open problems and identifying opportunities for future research. In this way, the paper invites the scientific community to contribute to research in TDL.

\care{\textbf{Why TDL?}
To articulate why TDL plays a critical role in the encoding, modeling and analysis of relational data, four practical advantages of TDL are presented. First, the topology of the underlying data space determines the choice of possible neural network architectures. 
Second, topological domains enable the modeling of data containing multi-way interactions (also known as higher-order relations). Third, TDL captures regularities inherent to manifolds, such as `remeshing symmetry'. Fourth, TDL captures topological equivariances in the data. In summary, TDL takes into account topological characteristics that appear in relational data, and therefore is a natural choice for various machine learning problems. Figure~\ref{fig:img1} shows a range of examples in which topological approaches to deep learning are useful and sometimes crucial.}

\care{\textbf{Choice of neural network architecture.}}
\citet{olah2014neural} contributed one of the first works on TDL, showing that the topology of the underlying space must be considered when choosing a neural network architecture.
In the early stages, the term `TDL' was often used to refer to the incorporation of features generated by persistent homology within the input pipeline of a deep neural network~\citep[DNN;][]{cang2017topologynet,hofer2017deep}. 
However, this paper uses the term `TDL' to refer to the collection of ideas and methods related to the use of topological concepts in deep learning. Following~\citet{Hensel21}, TDL methods can be used in an observational fashion, improving the understanding of existing deep learning models and their topological formulations, or in an interventional fashion, allowing deep learning architectures to treat data supported on higher-order domains, such as cell complexes~\citep{hajijcell,bodnar2022b,giusti2023cell}, simplicial complexes~\citep{bodnar2022a,schaub2022signal,Mitchell2023}, sheaves~\citep{hansen2019toward,Bodnar}, combinatorial complexes~\citep{hajij2023tdl}, and hypergraphs~\citep{feng2019hypergraph,kim2020hypergraph,bai2021hypergraph,behrouz2023catwalk}. 

\care{\textbf{Multi-way interactions.}}
In TDL, multi-way interactions between entities constitute useful features capable of embedding topological structure via deep learning algorithms. Higher-order relations capture long-distance or seemingly disparate connections in a system, providing scope for effective or robust message-passing schemes~\citep{hajijcell,roddenberry2021principled,bodnar2022a,hajij2023tdl, yang2023convolutional}.
\care{While graph-based equivariant neural networks have been developed to incorporate certain multi-way interactions~\citep{batatia2022,musaelian2023}, many types of multi-way interactions can arise in practice. TDL operates on various topological domains, such as simplicial and cellular complexes, and therefore provides a framework to model a multitude of possible types of multi-way interactions appearing in relational data.}

\care{\textbf{Regularities inherent to manifolds.}}
Beyond higher-order relations, the topological view may also capture the regularities inherent to manifolds. Examples include `remeshing symmetry' over manifolds, such as being invariant to different triangulations of a sphere or inducing similar behaviors at different meshing resolutions. These regularities would be perhaps not impossible, but quite challenging, to express using purely combinatorial constructs offered by graph representation learning (GRL), while TDL is a natural domain to think about such effects.

\care{\textbf{Topological equivariances.}}
\care{TDL is a natural approach to capture `topological equivariances'. For example, if a classification algorithm is meant to identify different knots, then it is useful to understand the stabilizer group involving isotopies of the complement. In general, GNNs and GDL are based on `standard groups'. For instance, GNNs adopt the permutation group, and applications of GDL in molecular modeling use subgroups of the Euclidean group, such as the special Euclidean group $\mbox{SE}(n)$ or the special orthogonal group $\mbox{SO}(n)$. TDL incoporates more complex homeomorphism groups that act on a space, while trying to preserve some embedding information. This embedding information can include, for example, the arrangement of critical points in Morse-Smale complexes~\citep{catanzaro2020} or the nesting of circles which leads to the concept of `tree of shapes' in image processing~\citep{caselles2009}. In particular, image segmentation provides an example of a topological structure that needs to be preserved under homeomorphisms of the domain.}

\textbf{Position.}
\care{\textbf{This position paper argues that TDL is the new frontier for relational learning. Relational data constitute a main modality of data emerging from natural and artificial systems~\citep{velickovic_everything_2023}. In such systems, sets of of objects are interconnected via binary or higher-order relations, and relational data encode features of these interconnections. TDL provides a framework for learning from binary or higher-order relational data.}}

\care{While TDL is pivotal in relational learning, it is more generally a valuable framework with many functions in the contemporary AI landscape. TDL informs the choice of neural network architecture subject to the topology of the underlying data space, captures regularities inherent to manifolds, and captures topological equivariances.} 

\care{Topological concepts} are expected to play a crucial role in articulating the position of TDL in relation to overlapping fields, such as GRL or GDL. Finding the conditions under which TDL becomes the go-to machine learning tool for practitioners is one of the main research problems in TDL. This paper builds on the engineering and mathematical knowledge of the community to identify possible research directions to solve this problem.

\textbf{Paper structure.}
The remainder of the paper is organized into several sections, each of which raises specific open problems and research directions for TDL. The ordering of the sections starts with more concrete application-centric topics and ends with issues related to the theoretical foundations of TDL, before concluding with some final remarks. An extensive literature review of TDL is provided for the interested reader in appendix~\ref{app:lit_review}.

\section{Examples, Datasets and Benchmarks} 

\care{This section highlights compelling applications and success stories of TDL in machine learning challenges to corroborate the advantages of TDL in various settings}. In addition, it proposes ways to enrich the range of datasets and benchmarks that can facilitate the evaluation of TDL methods.

\subsection{Examples}
\label{subsec:examples}

A natural use case of TDL involves attributed graphs, which combine structural information with feature information. Such graphs arise in numerous domains, for example, involving protein structures~\citep{xia2014persistent,sverrisson2021}, drug design~\citep{cang2017topologynet}, virus analysis~\citep{chen2022persistent}, or structural representations of molecules~\citep{jiang2021topological} and materials~\citep{reiser2022, townsend2020representation}. However, applications that require higher-order topological structures have been limited to intrinsically complex data, such as those arising in biological sciences~\citep{cang2018representability}. Perhaps the most compelling examples and applications of TDL which consistently demonstrate the relevant advantages of TDL over existing methods are the victories of TDL in the D3R Grand Challenges ~\citep{nguyen2019mathematical,nguyen2020mathdl}, the discovery of SARS-CoV-2 evolution mechanisms \citep{chen2020mutations,wang2021mechanisms}, and the successful forecasting of SARS-CoV-2 variants BA.2 \citep{chen2022omicron}, BA.4 and BA.5~\citep{chen2022persistent}  about two months in advance. 
More broadly, beyond graph-centric applications, other applications have benefited from a topological perspective of machine learning, including 2D shape analysis, unsupervised learning~\citep{Hofer19a}, and classification~\citep{Chen19a,Hofer20a,curry2023topologically}.

\begin{problem}
\label{prob:apps}
Compelling applications of TDL \care{have been developed, in which the topological properties of the data empirically demonstrate a competitive edge. However, a broader adaptation of TDL, which can revolutionize relational learning, has not yet taken place in real-world applications.}
\end{problem}

Several application areas are plausible candidates for TDL to shine, since their underlying domains give rise to topological structures. To encourage the development \care{and adaptation} of TDL \care{in practice}, the presence or emergence of \care{TDL applications in numerous} scientific disciplines is elaborated \care{further} in appendix~\ref{app:apps}.

\begin{direction}
\label{direction:apps}
Topological structures emerge in several scientific areas, including data compression, natural language processing (NLP), computer vision and computer graphics, chemistry, biological imaging, virus evolution, drug design, neuroscience, protein engineering, chip design, semantic communications, satellite imagery, and materials science. Synergies with researchers from these scientific disciplines are encouraged to develop real-world or impactful applications of TDL.
\end{direction}

\subsection{Datasets}

In addition to showcasing the practical benefits of TDL, applications can play an important role in the development and deployment of TDL models. In particular, standardized datasets derived from applications are instrumental in driving TDL research. The Open Graph Benchmark, 
a set of benchmark datasets, has been developed to facilitate reproducible graph machine learning research~\citep{hu2020,hu2021}. However, the community has not yet invested in resources to construct higher-order data and associated benchmarks because many graph methods do not take advantage of this information.

\begin{problem}
\label{prob:data}
There is a scarcity of higher-order data. One of the major milestones for the advancement of TDL is the prolific generation of public higher-order datasets.
\end{problem}

Higher-order data can be collected or synthesized. The TDL application areas of subsection~\ref{subsec:examples} are potential candidates to collect higher-order data. Conversely, a graph can be lifted to a higher-order domain~\citep{papillon2023}. Lifting procedures for graphs supply mechanisms for the generation of synthetic topological data. This is why a survey of lifting procedures and associated message-passing schemes would be useful for TDL. Moreover, a generalization of graph-rewiring approaches and analyses to higher-order topological structures would be a fruitful direction.

\begin{direction}
\label{direction:apps_and_lifting}
Developing applications of TDL can produce higher-order datasets that naturally arise from the underlying domain. A systematic assessment and generalization of graph-lifting and rewiring algorithms is a plausible path towards synthetic higher-order datasets.
\end{direction}

\subsection{Benchmarks}

The curation of a collection of higher-order datasets can pave the way for TDL benchmarks. Benchmarking is necessary for the comparative assessment and development of novel TDL architectures and associated learning algorithms.

\begin{problem}
Benchmark suites are needed to enable efficient and objective evaluation of novel TDL research.
\end{problem}

The design of open source and reproducible benchmark suites for TDL requires a minimal collection of higher-order benchmark datasets, as well as implementations of graph-lifting algorithms for generating synthetic datasets in higher-order domains. To ease user experience, a taxonomy of higher-order datasets is a recommended feature, organizing benchmarks, for example, by dataset size and type of learning task. Benchmark suites for TDL are expected to have a comprehensive set of performance metrics that extend beyond predictive performance. For example, stability metrics are relevant to TDL, since it is anticipated that exploitation of topological structure in data can improve stability comparably to GNNs.

\begin{direction}
\label{direction:benchmark}
The basic components of TDL benchmark suites include higher-order datasets, graph-lifting algorithms, and predictive and stability metrics.
\end{direction}

\section{Software}

There are several graph-based learning software packages, such as \texttt{NetworkX}~\citep{hagberg2008}, \texttt{KarateClub}~\citep{rozemberczki2020}, \texttt{PyG}~\citep{fey2019}, and \texttt{DGL}~\citep{wang2019}. \texttt{NetworkX} facilitates computations on graphs, and \texttt{KarateClub} implements algorithms for unsupervised learning on graph-structured data. \texttt{PyG} and \texttt{DGL} are two geometric deep learning packages for graphs.

To the best of the authors' knowledge, four software packages provide functionality on higher-order structures, namely \texttt{HyperNetX}~\citep{liu2021}, \texttt{XGI}~\citep{landry2023}, \texttt{DHG}~\citep{feng2019hypergraph}, and \texttt{TopoX}~\citep{hajij2024topox}. \texttt{HyperNetX} enables computations on hypergraphs, while \texttt{XGI} provides similar functionality on hypegraphs and simplicial complexes. \texttt{DHG} is a deep learning package for graphs and hypergraphs. \texttt{TopoX} is a suite of Python packages designed to compute and learn with topological neural networks. The suite consists of three packages, \texttt{TopoNetX}, \texttt{TopoEmbedX} and \texttt{TopoModelX}. \texttt{TopoNetX} supports computations on graphs and higher-order domains, including colored hypergraphs, simplicial complexes, cell complexes, path complexes and combinatorial complexes. \texttt{TopoEmbedX} provides methods to embed higher-order domains into Euclidean domains. \texttt{TopoModelX} implements the majority of topological neural networks surveyed in~\citet{papillon2023}.

\texttt{pytorch-topological}
combines several state-of-the-art packages for TDA, including \texttt{giotto-tda}~\citep{Tauzin21a} and \texttt{Ripser}~\citep{Bauer21a}, thus enabling the creation of topology-driven algorithms that work on point clouds or structured data such as images. Similarly, \texttt{torch-ph}~\citep{perez2021giotto}
and \texttt{TopologyLayer}~\citep{Gabrielsson20a} support persistent homology computations, as well as differentiating through these computations to facilitate the development of topology-informed loss functions. \texttt{GUDHI}~\citep{clement2014} provides a wider range of methods, including the optimization of functions based on persistent homology.

However, despite numerous theoretical advances in TDL, practical implementations are scarce due to the limited availability of easy-to-use software packages for deep learning on higher-order structures. TDL is a broad research area, so the contemporary TDL software landscape needs to be enriched with more functionality to cover additional higher-order domains \care{(including domains related to higher-order multiplex networks, and dynamic versions of existing higher-order domains)}, graph-lifting and rewiring algorithms, TDL models, and learning algorithms~\citep{alain2023gaussian}.

\begin{problem}
\label{prob:software}
Research experimentation and deployment of TDL models are often hindered by the limited availability of software. Software development for TDL is one of the most pressing open problems in accelerating the progress of theoretical and engineering research in TDL.
\end{problem}

To accelerate the development of TDL software, more human capital and financial investment are required. Since the currently available TDL software packages are open source, there are software engineers and researchers who volunteer to improve these packages. Coding challenges~\citep{papillon2023} help in this direction. However, more programmers are needed. Academic research grants are one funding avenue for computer scientists to contribute to the development of TDL software. To attract financial resources from investors, it is expected to demonstrate the value of TDL. In other words, solving open problem~\ref{prob:apps} (compelling examples of TDL) creates conditions to address open problem~\ref{prob:software} (resources for the development of TDL software).

Transformers are popular due to their easy implementation in graphics processing units. As existing TDL methods might be more challenging to implement than transformers, software users might resort to transformers equipped with some topological encoding. This latter possibility provides an alternative direction for TDL research. It becomes apparent that understanding computational trade-offs and hardware friendliness of TDL algorithms is essential for the development and broad adaptation of TDL software. Advances towards hardware-friendly TDL implementations may include, for example, graph rewiring~\citep{topping2022understanding,nguyen2023} and positional encoding~\citep{wang2022equivariant}. 

\begin{direction}
\label{direction:software}
The acquisition of more resources is required to advance TDL software development. Furthermore, research on the computational trade-offs and hardware friendliness of TDL algorithms might yield more accessible TDL software implementations.
\end{direction}

\section{Complexity and Scalability}

While the development of software for TDL is one problem, another imminent question is whether the benefits of TDL outweigh the resulting costs. 
\care{The innovative design of TDL architectures can make a difference in practical applications. In the context of such applications, computational complexity and scalability are major challenges that become particularly prominent in TDL due to the inclusion of higher-order data.}
This section discusses the complexity and scalability implications of TDL.

\subsection{Complexity}

TDL models increase space and time complexity in comparison to GNNs, since the former operate on domains with supplementary higher-order structure. Thus, the question becomes whether the increased complexity of TDL is worthwhile.

\begin{problem}
\label{prob:complexity}
A cost-benefit analysis framework for TDL has not been formalized. The \care{increased} time and space complexity of TDL \care{relatively to GDL and GRL (due to the inclusion of higher-order data)} must be \care{systematically} factored in to enable informed decision-making when it comes to choosing between TDL, GDL or GRL.
\end{problem}

An approach to understanding the role of complexity in TDL is to analyze the trade-off between complexity and performance as a function of higher-order structure. For example, the efficiency of a TDL algorithm may be evaluated by connecting the increase in predictive performance through the inclusion of higher-order features with the increase in the computational cost of neural network training.

Another approach to dissecting the notion of complexity in TDL is to study the propagation of topological information through neural network layers. For example,~\citet{naitzat2020} have demonstrated that neural networks transform topologically complicated input data into topologically simpler forms as they pass through the layers.\care{~\citet{petri2020on} have expanded the work of~\citet{naitzat2020} by showing that it is not necessarily the topology of the data that is simplified, but the topology of the decision boundary.}~\citet{wheeler2021} have studied how the persistence homology of data transforms as the data pass through successive layers of a deep neural network. While~\citet{naitzat2020},\care{~\citet{petri2020on} and}~\citet{wheeler2021} do not focus on input data supported on higher-order structures, such works provide plausible ways of characterizing topological complexity, for instance, via layer-wise transformations
of the underlying topology of the data.

\begin{direction}
\label{direction:complexity}
Complexity in TDL may be formalized in terms of performance-complexity trade-offs dependent on higher-order (algebraic) structure or from the angle of propagation of topological information through neural network layers.
\end{direction}

\subsection{Scalability}

Limited software availability (open problem~\ref{prob:software}) is not the only challenge in the development of TDL. Lack of scalability impedes progress in TDL. Besides, the multitude of higher-order domains confines the interoperability of scalable TDL models and learning algorithms. For instance, if a scalable TDL model is developed for simplicial complexes, it would not be applicable to other higher-order domains, such as cell complexes, hypergraphs or path complexes.

\begin{problem}
\label{prob:scalability}
Building scalable TDL models, which work across multiple higher-order domains, remains an open problem.
\end{problem}

One way of building resource-efficient TDL models might be to leverage more information from fewer data points or to develop optimization algorithms that converge provably faster on higher-order domains.
Another direction, which has seen improved performance in TDA applications, involves subsampling~\citep{Moor20a, wagner2021improving}.

Distributed training emerges as another plausible path to scalable TDL. Message-passing schemes on graphs naturally lead to distributed learning algorithms by letting `messages' travel through separate nodes. For example,~\citet{scardapane2021} have introduced a framework for distributed GNN training. One possibility is to extend the work of~\citet{scardapane2021} toward distributed TDL.

\begin{direction}
\label{direction:scalability}
Extending existing results in distributed GNN training may pave the way to scalable training of TDL models.
\end{direction}

\section{Explainability, Generalization and Fairness}
\label{sec:expl_general_fair}

This section focuses on the problems of explainability, generalization, and fairness.
\care{Such problems are common in machine learning and have been explored in other contexts, but how these can get translated in the context of TDL is largely unexplored. This section}
discusses how TDL can play an important role in solving these problems.

\subsection{Explainability and Generalization}

Understanding deep learning is a remarkably difficult task and is one of the main concerns of the community. One question is how topology can contribute to the understanding of deep learning, including explainability for neural network weights, decisions, training mechanisms, and characterization of generalization error.

\begin{problem}
\label{prob:explainability}
Harnessing topology to address questions related to explainability and generalization in deep learning delineates a family of open problems.
\end{problem}

Recent results have demonstrated that the topology of data, training trajectories, neural network weights and internal representations are related to generalization error in deep learning~\citep{rieck_neural_2019, Hofer20a, Birdal21, Andreeva23a, Dupuis23a}. Along these lines, a feasible research endeavor to predict the generalization error may involve training of TDL models on the computational graph, training trajectories of neural networks, or promoting desirable topological properties of internal representations that are linked
to generalization. The upshot of any advances along these directions would be to draw greater attention to TDL, given the potential impact of TDL on fundamental deep learning research.

\begin{direction}
\label{direction:explainability}
Training TDL models on the computational graph, controlling topological properties of weight trajectories traced out during stochastic optimization of neural networks, or appropriately enforcing desirable topological properties of a network's internal representation of the data may hold the keys to understand the puzzling generalization behavior in deep learning and to move a step closer to explainable AI.
\end{direction}

\subsection{Fairness}

In addition to explainability and generalization, there are other topics in GDL that have not yet been extended to TDL, such as model fairness~\citep{spinelli2022} and adversarial attacks~\citep{Zheng21a,ZhouZhouDing2023}. To this end, at least two issues need to be addressed, namely the definition of valid fairness metrics and the development of graph sampling or rewiring algorithms. 

\begin{problem}
\label{prob:fairness}
TDL model fairness has not been adequately studied.
\end{problem}

In a graph, fairness can be quantified, for example, using homophily, which counts the ratio of inter-class edges versus intra-class edges.~\citet{telyatnikov2023} have proposed a way to extend homophily measures to hypergraphs. By adapting the work of~\citet{telyatnikov2023}, the notion of homophily can be extended to higher-order features other than hyperedges. Bias in higher-order structures can be assessed by generalizing the assortative mixing coefficient~\citep{newman2003a} beyond graphs. Furthermore, dyadic fairness metrics for graphs, such as disparate impact~\citep{laclau2021} and statistical parity~\citep{rahman2019}, may be extended to polyadic fairness metrics for higher-order structures.

\begin{direction}
\label{direction:fairness}
To build fair TDL models, existing notions of fairness for graphs, such as homophily, assortative mixing and dyadic fairness, can be generalized for higher-order structures.
\end{direction}

\section{Theoretical Foundations}

This section is dedicated to the theoretical foundations of TDL. It focuses on open problems and associated research directions regarding the advantages of TDL, topological representation learning (TRL), and transformers in TDL.

\subsection{Advantages of Topological Deep Learning}


\textbf{Why is topology relevant in deep learning?}
A first step in answering this question \care{has been taken by developing} empirically verifiable examples in which TDL outperforms traditional learning methods in some respect (see also open problem~\ref{prob:apps}). Validating such possible empirical findings requires the establishment of theoretical answers. How can theory justify the relevance of topology in deep learning? Can theory prove (or disprove) that higher-order domains provide a richer representation than vector spaces?

\textbf{When do higher-order relations become useful?}
Higher-order relations may not benefit every application. Is it possible to discover the context in which higher-order relations become useful~\citep{battiston2020networks,benson2021higher,bick2021higher}? Is it possible to pinpoint higher-order relations that lead to improved downstream performance? Once discovered, can these relations be certified as semantically meaningful? Can physical or biological relevance be drawn from higher-order relations with visual tools or semantic metrics? Can higher-order relations help scientists understand phenomena that are currently obscure?

\textbf{\care{What are the advantages of TDL over GRL?}}
\care{One question is whether in certain cases higher-order relations are not required and the same performance might be achieved by stacking more layers in regular GNNs or by using graph transformer networks~\citep{yun2019}. This can potentially happen, but it is known that neural architecture design matters (especially in the low-data regime) with respect to capturing higher-order relations. For example, the work of~\citet{sanford2023representational} implies that existing transformers are efficient in modeling pairwise relations, but not three-way relations. Furthermore, TDL may help avoid some of the common pathologies that arise with deeper GNNs, such as vanishing gradients and degradation.}

\begin{problem}
\label{prob:tdl_advantages}
Theoretical foundations have not yet been adequately laid to consolidate the relative advantages of TDL. More theoretical research is needed to shed light on the relevance of topology in deep learning and contextual understanding of downstream performance gained by higher-order relations.
\end{problem}

\textbf{Expressivity proofs for TDL.}
For some GNNs and datasets, it is known both theoretically and empirically that persistent homology outperforms GNNs~\citep{Horn22a}. There are graphs with two connected components or two cycles that cannot be distinguished by the ordinary Weisfeiler-Lehman test for graph isomorphism and, thus, also not by message-passing GNNs. Furthermore, the increased expressivity of higher-order message passing has been proven~\citep{bodnar2021weisfeiler}. Expressivity proofs provide a mechanism to unravel the relevance of topology in deep learning. For example, conditions may be sought under which DNNs or GNNs can or cannot learn certain topological invariants.

\textbf{Studying properties of TDL.}
One way to evaluate the relevance of the topology in TDL is to study the properties of TDL. This raises the question of which properties of TDL to focus on. For instance, investigating over-smoothing and over-squashing in deep TDL models may provide useful insights for the relative advantages of TDL over GNNs. Interpreting the spectral properties of Hodge Laplacians may illuminate how message passing benefits from topological structures. It is also worth examining whether neural network parameter priors based on higher-order structures yield any gains in reliability or uncertainty quantification.

\textbf{Generative TDL.}
Generative modeling aims to learn the distribution of training samples by generating new instances. While image generation is now well-studied ~\citep{rombach2022highresolution}, generating real-world data supported on non-Euclidean or mixed domains presents an unresolved challenge. This has led to research on the generation of 3D point clouds~\citep{zhao20193d,zeng2022lion}, meshes~\citep{Liu2023MeshDiffusion,poole2022dreamfusion}, and graphs~\citep{guo2022systematic,vignac_digress_2023}. Yet, there is no unifying representation that can generate all these 3D representations. This is one of the premises of TDL.
In the field of structural biology, the generation of molecules and proteins is currently approached as graph generation~\citep{vignac_digress_2023,hoogeboom2022equivariant,ingraham2019generative,zhang_survey_2023} or by generative modeling on 3D rigid bodies~\citep{yim2023se}. Generative modeling for graphs generally involves creating new graphs from scratch. However, for many applications where data have inherent geometric structures, relying solely on graphs, as done in~\citet{velickovic_everything_2023}, can potentially omit useful higher-order information. Recent studies have shown that considering higher-order relations yields more accurate representations of both scene graphs and molecular structures~\citep{zhan2022hyper,watson2023novo}.
In addition to 2D and 3D computer vision, diffusion models have become the state-of-the-art in graph-based molecular generation~\citep{xu2022geodiff,vignac_digress_2023,haefeli2022diffusion,jing_torsional_2023}. In particular,~\citet{jing_torsional_2023} have introduced torsional diffusion, a hybrid method that uses topological principles to diffuse in a toroidal space, narrowing the search space and improving both model performance and inference speed. This approach demonstrates how topology can enhance existing generative methods. Despite these advances, the development of topological diffusion and flow-matching models remains an uncharted territory.
Data characterized by higher-order structures align naturally with TDL. In some cases, such structures are not inherently present, but can be inferred~\citep{tang2023hypergraph,tang2023hypergraph2}, raising questions about the nature and implications of generating higher-order features absent from the original data. The assignment of specific semantic meanings and structures to these generated forms is an ongoing research challenge, paving the way for more interpretable TDL. This area of research, both recent and progressive, holds the potential to advance generative TDL.

\textbf{Sheaf theory for TDL.}
There has been some work on categorical machine learning~\citep{fong2021}, as well as TDA to study activation layers~\citep{naitzat2020}, but there has not been much work connecting the two. Category theory offers rich structures for knowledge representation, but these structures are typically quite rigid. Persistent homology is less committal and allows algebraic structure to come into existence as a function of some metrized parameter space. Sheaf theory can serve as a bridge between category theory and persistent homology in the context of TDL. For example, a combination of applied category theory and TDA may lead to algorithms that automatically summarize how neural networks produce their output, making deep learning more understandable to humans. Moreover,~\citet{fong2021} have introduced a theoretical sheaf approach to detect local merging relations in digital images, and~\citet{bodnar2022b} have developed neural sheaf diffusions to identify topological data features that are important for real-world problems. Persistent sheaf Laplacians allow the incorporation of physical laws in the topological representation of data~\citep{wei2021persistent}. The neural sheaf diffusion models of~\citet{bodnar2022b} address some limitations of classical graph diffusion equations and the corresponding GNNs, providing ideas that can be extended to push the limits of TDL through algebraic topology.

\begin{direction}
\label{direction:tdl_advantages}
Expressivity proofs may offer a comparative characterization of topological invariance in GNNs and TDL models. Properties of TDL models, such as over-smoothing and over-squashing, and spectral properties of Hodge Laplacians may help demystify the relevance of topology in TDL. Sheaf theory for generative TDL is a plausible research avenue to demonstrate how twisting of feature spaces in a GNN affects expressivity and detects inconsistent logic present in large language models.
\end{direction}

\subsection{Topological Representation Learning}

There are several possible interpretations of the umbrella term TRL. \care{As a high-level definition, TRL is concerned with automatically learning and exploiting multiscale topological descriptions of the data and of the neural network's internal embeddings, which scale efficiently to larger dimensions and which provide benefits in efficiency and transparency compared to the equivalent graph-based representations.}

\care{One question in TRL is how} to develop training strategies that enable a neural network to elicit higher-order information from the data for the task at hand. An immediate follow-up question in this context is whether this uncovered higher-order structure is meaningful in any semantically valid sense. In the context of GNNs, it is, for example, unclear whether such an uncovered higher-order structure might not be equivalent to the message-propagation phase in the spirit of~\citet{velivckovic2022}. However, even if this is true, it might still be possible to analyze these structures to improve the explainability and transparency of a neural network.

\care{Another question in TRL is how} to develop methods that promote or penalize certain topological properties of a neural network's internal representation of the data or its output. On the one hand, such properties could be informed by a priori knowledge or corresponding properties in the input domain, as successfully demonstrated in image segmentation problems~\citep{Hu21a,Gupta23a}, learning representations to synthesize new shapes~\citep{Waibel22a}, using random walks on complexes to obtain cell embeddings~\citep{billings2019simplex2vec,schaub2020random}, or preserving topological properties while training autoencoders~\citep{Moor20a,hajijcell,trofimov2023learning}.
On the other hand, topological characteristics of representations can be guided by establishing provably beneficial properties for generalization~\citep{Hofer19a, Hofer20a} \care{or can be used for comparing neural network representations~\citep{barannikov22a}.}
Another facet of this interpretation of representation learning involves the analysis and modification of the decision boundaries of classifiers. Here,~\citet{Chen19a} have shown that regularizers based on persistent homology lead to better generalization and higher predictive performance.

\begin{problem}
\label{prob:representation_learning}
It remains unclear how to automate the learning of topological representations in TDL, that is, how to obtain semantically meaningful topological information from the data during neural network training.
It is an open problem as to what properties of representations to promote or penalize when interested in generalization questions in TDL. Further, it is often far from straightforward how to appropriately encode desirable properties of the input space, not least due to the complexity and possibly high-dimensionality of the data.
\end{problem}

The explainability of predictions made by GNNs~\citep{luo2020,lucic2022} is complicated by the fact that even a small subgraph is difficult for a human to visualize and grasp quickly. Such complications are amplified in TDL, since visualization and human apprehension of higher-order domains is virtually nonexistent to date. Nevertheless, information in a higher-order domain can be summarized more flexibly by a smaller set of cells or hyperedges, leading to potentially interpretable structure. In other words, the richer structure attained by higher-order relations may enable more effective representation learning. For example, the work of~\citet{battiloro2023} on latent topological inference (LTI) has introduced a trainable method to identify higher-order interactions between entities, combining advances in TDL and differentiable sampling. Building upon LTI might lead to a general framework for TRL on arbitrary higher-order domains. Building on the graph-based representation learning approach of~\citet{hamilton2017representation}, the work of~\citet{hajij2022simplicial, hajij2023tdl} has introduced a TRL framework for neighbourhood-based methods, which provides scope for topological neural network architecture search. 

\begin{direction}
\label{direction:representation_learning}
Higher-order relations, encoded by cells or hyperdeges, offer candidate components for effective TRL. Latent topological inference outlines a tentative methodological path toward generalized representation learning for TDL. Existing work on TRL provides a means to develop topological neural network architecture search algorithms.
\end{direction}

\subsection{Transformers in Topological Deep Learning}
Looking for a `transformer architecture' for graphs has led to graph transformer models~\citep{dwivedi2020generalization,min2022transformer}, along with the first `foundation models' for graph datasets~\citep{liu2023towards}, which can be transferred (either in a zero-shot fashion or with fine-tuning) to smaller tasks. It is not yet known what the `transformer architecture' of TDL models is, if it exists. It is expected that over the next years new transformer models for topological domains may be defined, along with algorithms such as diffusion generative algorithms or structured state-space models for cell and simplicial complexes.

Conversely, it is interesting to explore whether higher-order variants of transformers can be used to improve tasks in mainstream domains, if computational bottlenecks can be solved. For example, is there any meaningful latent information embedded in text or images that can be learned and leveraged by a higher-order transformer?

\begin{problem}
\label{prob:tdl_transformers}
Developing a transformer architecture for TDL models would lay a unified foundation for TDL across different higher-order domains. A follow-up question would be the applicability of such a `topological transformer' to text, audio, or imaging data.
\end{problem}

Advective diffusion transformers have been introduced by~\citet{wu2023} to generalize GNNs in the presence of varying graph topologies. A first step in extending the work of~\citet{wu2023} might be to consider diffusion transformers inspired by diffusions on hypergraphs~\citep{prokopchik2022,wang2022} or heat diffusion on simplicial complexes~\citep{aktas2021}.

\begin{direction}
\label{directions:tdl_transformers}
A first milestone for a transformer architecture for TDL might be a transformer architecture that is specific to a given higher-order domain. For instance, a transformer might be inspired by existing diffusions on hypergraphs or simplicial complexes.
\end{direction}

\section{Final Remarks}

To accelerate the transition of TDL from theoretical constructs to real-world applications, the research community is invited to adopt a multidisciplinary approach rooted in innovation, collaboration, and open science. Such an approach necessitates the convergence of diverse fields, bringing together experts in mathematics, computer science, and machine learning, to foster a rich cross-pollinating environment. By encouraging open dialogue and the exchange of ideas, TDL can evolve rapidly, breaking through current barriers, and sparking groundbreaking advancements.

One way to facilitate this interdisciplinary approach is through the organization of TDL workshops and conferences. These gatherings are the key to creating platforms for knowledge sharing, discussion, and the initiation of collaborative ventures for TDL. In these events, it is important to adopt open science principles.


This paper can help guide collaborative efforts and research discussions on TDL. It highlights key challenges and outlines actionable research strategies that can pave the way for concerted and effective efforts in the field. By openly sharing ideas and research directions in this paper, the goal is to support collaborative efforts, promote transparency, and accelerate innovation in TDL. This culture of openness allows researchers from various fields to contribute and benefit from the advances in TDL. 





\section*{Acknowledgements}
This work is partially supported by
DOE grant DE-SC0023157 (BW),
NIH grants R01AI164266 (GWW) and
R35GM148196 (GWW), and
NSF grants 2134241 (NM, MH),
DMS-2134223 (BW),
IIS-2205418 (BW),
CCF-2112665 (YW),
CCF-2310411 (YW),
DMS2052983 (GWW)
and IIS-1900473 (GWW).
This work was partially funded by the Italian NRRP (PE00000001 - program `RESTART'), and by the 6G-GOALS project under the 6G SNS-JU Horizon program, n.101139232 (PDL).
VM is partially funded by NSF DMS No 2012609, US Army Research Lab Contract No W911NF-21-2-0186, US Army Research Lab Contract No W911NF-22-2-0143, US Army Research Office No W911NF-21-1-0094, and The University of Tennessee Materials Research Science \& Engineering Center – The Center for Advanced Materials and Manufacturing – NSF DMR No. 2309083.
BR is supported by the Bavarian state government with
funds from the Hightech Agenda Bavaria.
SS is partly funded by the Sapienza grant RM1221816BD028D6 (DESMOS).
MTS acknowledges funding by the European Union (ERC, HIGH-HOPeS, 101039827). Views and opinions expressed are however those of the author(s) only and do not necessarily reflect those of the European Union or the European Research Council Executive Agency. Neither the European Union nor the granting authority can be held responsible for them.

The authors thank Cristian Bodnar for his input on the theoretical foundations of TDL, and the ICML reviewers, Thomas Kipf and Charles Blundell for their reviews and feedback.






\bibliographystyle{icml2024}

\newpage
\appendix
\onecolumn

\section{Literature Review}
\label{app:lit_review}

This appendix provides a brief literature review focused on TDL, while also broadly discussing the role of topology in machine learning. Three main themes are considered, namely higher-order deep learning models, research at the intersection of the topology of data and neural networks, and research that connects TDA with neural networks.

\subsection{Higher-Order Deep Learning Models}

Recent years have witnessed a growing interest in higher-order networks~\citep{mendel1991tutorial,battiston2020networks,bick2021higher} due to their effective ability to capture higher-order interactions. In signal processing and deep learning, various approaches, such as Hodge-theoretic methods, message-passing schemes, and skip connections, have been developed for higher-order deep learning models.

The use of Hodge Laplacians for data analysis has been investigated by~\citet{jiang2011statistical, lim2020hodge} and has been extended to a signal processing context, for example, in \citet{schaub2021signal,sardellitti2021topological,roddenberry2021signal} for simplicial and cellular complexes. Also, edge-centric convolutional neural networks (CNNs) leveraging the 1-Hodge Laplacian operator for linear filtering have been defined~\citep{roddenberry2019hodgenet,barbarossa2018learning,schaub2018denoising,barbarossa2020topological,barbarossa2020topologicalmag,schaub2021signal}.

Convolutional operators and message-passing algorithms for higher-order neural networks have been developed. For hypergraphs, a convolutional operator has been proposed~\citep{jiang2019dynamic,feng2019hypergraph,arya2018exploiting}, and has been further investigated in~\citet{wu2022hypergraph,bai2021multi,jiang2019dynamic,bai2021hypergraph,gao2020hypergraph,giusti2023cell,gong2023generative}. In addition, a unified framework for learning on graphs and hypergraphs has been developed~\citep{huang2021unignn}, and general hypergraph neural networks have been introduced~\citep{gao2022hgnn}.

In the context of message passing on complexes, a general message-passing scheme on topological spaces has been developed by~\citet{hajijcell}, encompassing previous proposals~\citep{gilmer2017neural,ebli2020simplicial,bunch2020simplicial,hayhoe2022stable} and employing various topological neighborhood aggregation schemes. The expressive power of simplicial and cellular message passing neural networks has been studied by~\citet{bodnar2022a} and~\citet{bodnar2022b}, respectively. The work of~\citet{hajij2023tdl} provides the mathematical blueprint of topological deep learning, unifying existing deep learning architectures with a common mathematical language.~\citet{mathilde2023} has conducted a review of message-passing topological neural network architectures, employing foundational concepts from~\citet{hajij2023tdl}. This survey highlights key challenges and promising directions for future research in the field.  
\care{~\citet{zia2024topological} have carried out a review of TDL, focusing primarily on how the use of TDA techniques has evolved over time to support deep learning frameworks.}
Recurrent simplicial neural networks for trajectory prediction have been introduced~\citep{Mitchell2023}, and a multi-signal approach on higher-order networks utilizing the Dirac operator has been developed~\citep{calmon2022higher,hajij2023combinatorial}. 

While the dominant trend of topological neural networks represents a special case of higher-order message passing~\citep{hajij2023tdl}, recently, several notable topological neural network methods have been introduced that do not rely on this paradigm. For example,~\citet{Maggs23a} have introduced neural $k$-forms, providing a novel approach to operate on geometric simplicial complexes and graphs without relying on traditional message passing. In addition,~\citet{ramamurthy2023topo} have introduced an MLP-based simplicial neural network algorithm to learn the representation of elements in a simplicial complex without explicitly relying on message passing, ensuring fast inference time and high robustness against the lack of connectivity information during inference.

To facilitate the training of higher-order DNNs, a generalization of skip connections to simplicial complexes has been proposed~\citep{hajij2022high}. Additionally, a higher-order GNN considering higher-order graph structures at multiple scales has been presented~\citep{morris2019weisfeiler}.

Most attention-based models are designed primarily for graphs, with some recent exceptions that have been introduced in higher-order domains~\citep{bai2021hypergraph,kim2020hypergraph,georgiev2022heat,giusti2022simplicial,goh2022simplicial,giusti2023cell}. For instance,~\citet{goh2022simplicial} have proposed a generalization of the graph attention model of~\citet{velickovic2017graph}, while~\citet{giusti2022simplicial} have introduced an attention model for simplicial complexes based on Hodge decomposition, similar to~\citet{roddenberry2021principled}. Furthermore,~\citet{giusti2022simplicial} have introduced a simplicial network architecture leveraging masked self-attention layers. Expanding this work,~\citet{battiloro2023generalized} have introduced a generalized attention architecture that integrates the interplay among simplices of different orders through the Dirac operator and its Dirac decomposition, adding a layer of complexity and depth to the understanding of attention mechanisms. Attention on cell complexes has been introduced by~\citet{giusti2023cell} exploiting higher-order topological information through feature lifting and attention mechanisms over lower and upper neighborhoods. In a related work, a generalized attention mechanism on combinatorial complexes has been introduced by~\citet{hajij2023tdl}. Hypergraph attention models in~\citet{bai2021hypergraph,kim2020hypergraph} provide alternative generalizations of the graph attention model by~\citet{velickovic2017graph}. These models collectively represent a generalization of the traditional graph-based attention framework to more complex and higher-dimensional structures.

Finally, tangent neural networks that operate over the tangent bundle of Riemann manifolds have recently been introduced by~\citet{battiloro2023tangent}.~\citet{battiloro2023} have extracted meaningful latent topology information tailored to specific tasks, providing a versatile approach for topology-aware neural networks.

\subsection{Topology of Data and Neural Networks}

Interpreting Euclidean data as a sample set from a topological space has a long history~\citep{Carlsson2009}. Some early points on the interplay between topology and neural networks appeared in a blog of~\citet{olah2014neural}. In particular,~\citet{olah2014neural} has performed topological experiments that illustrate the importance of considering the topology of the underlying data when choosing a neural network. 
Approximately at the same time,~\citet{Bianchini14a} have published their seminal work related to the expressivity of DNNs, later extended by~\citet{Sun16a}. In~\citet{naitzat2020}, the activations of a binary classification neural network have been considered as point clouds on which the layer functions of the neural network act. The topologies of these activations have been studied using homological tools such as persistent homology~\citep{EdelsbrunnerHarer2010}.~\citet{rieck_neural_2019} have studied the topology of individual layers in MLPs, developing neural persistence as a way to describe deep learning phenomena based on topological concepts. Some caveats of this approach have recently been observed and rectified by~\citet{Girrbach23a}, leaving the open question of how to effectively use topology to describe neural network training.~\citet{hajij2021tdl} have used topological notions to provide insights into the supervised classification problem in the context of neural networks. Similarly,~\citet{Oballe2022,Love2023} have used topological properties of data to infer deep learning architectures.

\subsection{Topological Data Analysis and Neural Networks}

Although an old field, algebraic topology~\citep{hatcher2005algebraic}  has only recently begun to solve problems in various domains. For example, persistent homology~\citep{EdelsbrunnerHarer2010} succeeded in solving complex data problems~\citep{attene2003shape, bajaj1997contour, boyell1963hybrid, carr2004simplifying, curto2017can, DabaghianMemoliFrank2012, giusti2016two, kweon1994extracting, LeeChungKang2011b, lum2013extracting, nicolau2011topology, rosen2017using}. Deep learning models rooted in topology have left their imprint in various domains, including topological data signatures~\citep{biasotti2008describing, carlsson2005persistence, rieck2015persistent}, neuroscience~\citep{curto2017can}, bioscience~\citep{ topaz2015topological} time series forecasting~\citep{zeng_topological_2021}, Trojan detection~\citep{hu_trigger_2022}, image segmentation~\citep{Hu19a}, 3D reconstruction~\citep{Waibel22a}, and time-varying setups~\citep{Rieck20}. TDA~\citep{EdelsbrunnerHarer2010,Carlsson2009,ghrist2014elementary,DW22,Love2023} has relied on topological tools to analyze data and generate machine learning algorithms. 

In~\citet{rieck_neural_2019}, the analysis of neural networks and their generalization performance has been approached through persistent homology, providing valuable observational insights. Building on this work,~\citet{Hofer20} have introduced a topology-based graph and simplicial complex readout function, paving the way for a deeper understanding of data representations. In pursuit of efficient representation learning,~\citet{Moor20a} have focused on topology-based methods, particularly emphasizing the use of the 1-skeleton for performance reasons while actively working to improve the speed of simplicial complexes.~\citet{Horoi22a} have adopted topological concepts to analyze neural network behaviors, providing observational insights into the learning process. Moving toward end-to-end learning,~\citet{Horn22a} have introduced a novel perspective, incorporating topological information into graph and simplicial complex learning. The work of~\citet{Horn22a} breaks free from the confines of traditional graph isomorphism tests, enhancing the understanding of complex learning structures. In the context of 2D-to-3D image reconstruction,~\citet{Waibel22a} have demonstrated how topological information, specifically cubical complexes in 3D, leads to more efficient models and higher quality reconstruction results. Exploring has continued in~\citet{Nadimpalli23a}, focusing on quantized and approximated Euler characteristic transforms for image reconstruction, contributing to the arsenal of tools to enhance reconstruction processes. Taking a singular approach,~\citet{vonRohrscheidt23a} have formulated a method for detecting multi-scale singularities in data using local persistent homology. This observational study has linked singularity information with generalization performance, adding depth to understanding the difficulty of classification tasks. In pursuit of high-performance representation learning, \citet{Roell23a} have introduced a fully differentiable variant of the Euler characteristic transform, facilitating tasks on graphs and combinatorial complexes. The use of curvature-based topological information has been showcased by~\citet{Southern23a}, offering a perspective on the evaluation of generative graph models. The interplay between probabilistic machine learning and topology has recently been investigated in several works. Specifically,~\citet{Maroulas2019,Maroulas2020b,Oballe2022b,Papamarkou2022} have developed sampling techniques for distributions of topological summaries either in a non-parametric way or from a Bayesian perspective. Using variations of Cech and Vietoris-Rips, the topology of decision boundaries of neural networks has been studied by~\citet{ramamurthy2019topological} to quantify the complexity of DNNs and to enable the matching of datasets to pre-trained models.

\section{Application Areas for TDL}
\label{app:apps}

This appendix highlights numerous domains of application in which TDL has shown or may show its potential. In particular, TDL is a promising tool for data compression, NLP, computer vision and computer graphics, quantum TDL, chemistry, biological imaging, virus evolution, drug design, neuroscience, protein engineering, chip design, semantic communications, satellite imagery, and materials science.

\textbf{Data compression.}
TDL has the potential to compress relational data by exploiting multi-way correlations between elements to obtain effective lower-dimensional representations. As an example,~\citet{bernardez2023topological} have proposed a TDL-based approach for compressing graph signals, demonstrating superior performance compared to both GNNs and feed-forward architectures when applied to two real-world internet datasets. Additionally,~\citet{battiloro2023parametric} have introduced a topological dictionary learning algorithm for sparse signal representation over regular cell complexes. This method has demonstrated advantages in edge flow compression tasks, further illustrating the potential of topological methods compared to graph-based methods. These studies suggest that TDL holds great promise for advancing data compression techniques by leveraging its ability to capture complex correlations in relational data.

\textbf{Natural language processing.}
Deep learning has recently made impressive advances in the field of NLP, a rich and less explored domain to apply TDL. Specifically, contextualized word embeddings from large language models have revolutionized tasks such as text generation, sentiment analysis, and machine translation. Exploring the topological structures of these general-purpose learned embeddings can help NLP practitioners unlock the hidden structures of their models. TDL could be crucial to understanding how these models organize hierarchical class knowledge across neural network layers~\citep{PurvineBrownJefferson2023}. TDL can also probe the lexical, syntactic and semantic regularities of the embedding space~\citep{hajij2021data,RathoreZhouSrikumar2023}, and estimate its intrinsic dimensionality~\citep{Tulchinskii23a}, thus providing insights into model effectiveness. Additionally, the scope of TDL extends beyond word embeddings. It can be used to compare sentence embeddings from language models~\citep{Meirom22a}, study the parameter space of neural networks~\citep{GabellaAfamboEbli2019}, and investigate deep adversarial training~\citep{PerezReinauer2022,ZhouZhouDing2023}. This can help NLP practitioners understand the complex structures of their models and further advance language processing technologies.

\textbf{Computer vision and computer graphics.}
Several recent TDL works focus on learning directly on topological spaces. In the domains of vision and graphics, these spaces often manifest naturally as 3D point clouds, graphs, or meshes, either acquired in the real world through 3D sensors or designed in computer-aided design software. To date, the field has distinctly differentiated between these data representations, with the optimal choice often remaining elusive and highly dependent on the specific application~\citep{ahmed2018survey,shi2022deep}. The lack of a unifying representation has led vision and graphics researchers to develop specialized architectures for different data types. Examples include PointNet for point clouds~\citep{qi2017pointnet}, MeshCNN for meshes~\citep{hanocka2019meshcnn} and 3D GCNs for spatial graphs~\citep{lin2020convolution}.
Limited efforts have been made to consider a unified domain approach.~\citet{jiang2019ddsl} have attempted to bridge this gap via simplex mesh-based geometry representations, achieving success in shape optimization and segmentation on real-world datasets.~\citet{hajij2022simplicial} have employed simplicial complexes in representation learning, although their evaluation has been confined to meshes. The concept of combinatorial complexes~\citep{hajij2023tdl} has offered a unifying view and has been tested on mesh, point cloud and graph segmentation.

\textbf{Quantum TDL.}
Quantum neural networks (QNNs) are promising architectures for handling various types of datasets, which have the potential for exponential speed-up compared to classical neural network algorithms. Drawing inspiration from the success of GDL, a QNN has been introduced by~\citet{verdon2019quantum}, allowing for both quantum inference and classical probabilistic inference on data characterized by a graph-geometric structure. The computation of topological features often presents a formidable challenge, necessitating the subsampling of underlying data. A breakthrough in this domain comes from the work of~\citet{Ameneyro2022b}, which has introduced an efficient quantum computation method for persistent Betti numbers, using a persistent Dirac operator, whose square yields the persistent combinatorial Laplacian~\citep{wang2020persistent}. The approach of~\citet{Ameneyro2022b} unveils the underlying persistent Betti numbers that capture the enduring features of the data. To enhance its applicability to time series data,~\citet{Ameneyro2022} have introduced a quantum Takens delay embedding algorithm, which transforms time series into point clouds within a quantum framework. The work of~\citet{Ameneyro2022} broadens algorithmic utility and shows adaptability in various data structures. While existing QNNs do not explicitly cater to data defined over topological spaces, and overall quantum topological methods are in their infancy, the advent of TDL suggests a fertile research direction in this realm. Indeed, the emerging field of TDL may pave the way for tailored QNNs designed to process data within topological spaces, or develop QNN architectures that embody topological properties in their structures.

\textbf{Chemistry.} TDL can offer significant benefits to chemistry and molecular applications by providing a flexible framework that can effectively capture complex molecular structures, long-range dependencies, and higher-order interactions. As discussed in~\citet{jiang2021topological,Grier2023}, capturing topological properties offers a promising way to understand the underlying structure of a molecule. For example,~\citet{jiang2021topological} have demonstrated the power of methods based on algebraic topology in constructing unique representations of crystalline compounds. These methods not only capture intricate pairwise and many-body interactions, but also reveal the topological structure within groups of atoms at various scales. This approach, which goes beyond a molecular graph representation, has been successfully applied in molecular discovery, as presented in~\citet{townsend2020representation}, helping to uncover novel molecular structures and properties. Furthermore, TDL has been shown to be valuable for assessing and predicting the safety and potential risks associated with various chemical compounds~\citep{wu2018quantitative}. Another molecular application of TDL has been explored by~\citet{bodnar2021weisfeiler}, who have introduced a neural network based on a cell complex to predict molecular properties. The proposed network has provided enhanced expressivity, principled modeling of higher-order signals, and improved distance compression, leading to the achievement of state-of-the-art results in various synthetic and real-world molecular benchmarks. These studies suggest that topology plays a crucial role in supramolecular chemistry, particularly in the study of spatially organized multimolecular complexes characterized by diverse bonding patterns, such as mechanically interlocked and interpenetrated networks, thus highlighting the growing significance of TDL in the advancement of molecular and chemistry-related research. 

\textbf{Biological imaging.} Capturing 3D configurations of minuscule biological entities, such as proteins, presents a significant imaging challenge. A pivotal advancement in this area has been made through cryo-electron microscopy (cryo-EM), which has notably transformed structural biology. This technique enables the imaging of molecules within a solution, achieving a level of detail that was previously unattainable. The primary purpose of using cryo-EM is to reconstruct the 3D shape of biomolecules from the 2D images it produces. However, this reconstruction task is complex, as the initial images obtained from cryo-EM are noisy 2D projections of the underlying 3D biomolecules. A promising approach to enhance the accuracy of this reconstruction is the application of robust topological or geometric principles, acting as guides for what the 3D structure of the biomolecule might plausibly be. For example, using a known 3D structure of the protein as a reference, conceptualized as a cellular complex with nodes symbolizing the residues, edges representing covalent bonds, and faces indicating the rings, can significantly aid in the reconstruction process. Any further analysis or processing of these reconstructed structures can benefit from TDL, leveraging its capabilities to refine and understand complex biomolecular shapes derived from cryo-EM~\citep{xia2015persistent,biswas2016effective,zhao2020rham}. In addition, TDL has shown to be useful in scenarios involving high intra-subject variance, such as in magnetic resonance imaging (MRI). It facilitates the comparison and analysis of such complex datasets, thereby improving the effectiveness of learning tasks in these areas. The incorporation of topological structures as a means of regularizing various tasks has led to notable advances in areas such as image segmentation~\citep{Hu19a,Hu21a,Clough22a,Gupta22a} and image-based shape reconstruction~\citep{Waibel22a}. The application of topological concepts as a prior in these tasks has contributed to improved inference. These studies highlight the potential of TDL in biomedical imaging applications.

\textbf{Virus evolution.} During the COVID-19 pandemic, understanding the evolutionary trend of the SARS-CoV-2 virus was a great challenge. TDL has played a vital role in unraveling the evolutionary mechanisms of SARS-CoV-2, including the natural selection process that strengthens the infectivity, as elucidated in~\citet{chen2020mutations}, and the development of antibody-resistant spike mutations, as explained in~\citet{wang2021mechanisms}. This mechanistic understanding has been instrumental in forecasting the emergence of dominant variants such as BA.2~\citep{chen2022omicron} and BA.4, BA.5~\citep{chen2022persistent} approximately two months in advance. The contribution of TDL to understanding these mutations and their implications has been indispensable to addressing the challenges posed by the rapidly evolving SARS-CoV-2 virus. It exemplifies how TDL can be used effectively to analyze virus evolution and adapt to emerging threats.

\textbf{Drug design.} In life sciences, the development of effective drugs to combat diseases is of utmost importance. The central goals of molecular bioscience and biophysics are to decipher the molecular complexities of human diseases and to create drugs that can mitigate these diseases without causing harmful side effects. A crucial aspect of drug design and discovery is the ability to predict how a given molecule will interact with biomolecules such as proteins or DNA. This interaction is key to determining whether a drug can activate or inhibit the function of a biomolecule, leading to therapeutic benefits. The D3R Grand Challenges, an annual global competition in computer-aided drug design, served as a platform for testing and evaluating various methodologies from physical, chemical, statistical, and computer science perspectives in terms of scoring, ranking, and docking capabilities in drug design. In this challenge, TDL was a winner, outperforming other techniques in accurately predicting molecular interactions~\citep{nguyen2019mathematical,nguyen2020mathdl}.  
The effectiveness of TDL is particularly notable in drug discovery and other applications involving molecular structures, given the intricate balance between geometry (the physical space of molecules) and topology (the inherent connectivity of molecules). Benchmark studies have demonstrated that both dynamic TDL methods~\citep{Horn22a} and static topological features~\citep{Rieck19b} can result in enhanced predictive performance. This positions TDL as a promising area for future research and development in drug discovery. 

\textbf{Neuroscience.}
Exploring neuroscience through a topology-based lens proves instrumental in tackling the substantial intra-subject variance inherent in brain imaging modalities. The work of~\citet{Rieck20} has showcased the efficacy of static topological features in functional MRI (fMRI) data analysis and prediction tasks. The neural manifold hypothesis, as stated in~\citet{bengio2013representation}, posits that neuronal activity forms a low-dimensional manifold reflecting the structure of encoded task variables. Notably, the topology of the neural manifold aligns intricately with the task's topology. Examples include the topological ring in neural circuits that encode the direction of an animal's head and the torus in the grid cell circuit for spatial navigation~\citep{chaudhuri2019intrinsic, gardner2022toroidal}. TDL, with its application to reveal the scientific meaning behind neural code topology, emerges as a crucial tool. Learning the topology serves as a fundamental first step for subsequent analysis, including geometric analysis~\citep{acosta2023quantifying}. An intriguing facet of spatial information encoding involves grid cells, layered neurons facilitating environment-based navigation.~\citet{Mitchell2023} have captured the firing structure as simplicial complexes, decoding grid cell data by developing a recurrent neural network based on simplicial complexes. Furthermore, studies by~\citet{Nasrin2019} and~\citet{Maroulas2019} have addressed brain wave data stochasticity by detecting topological fingerprints in electroencephalograms and estimating the topological noise in such fingerprints. The comprehension, representation, and decoding of neural structures benefit from models with higher-order connectivity, making TDL integral to understanding intricate patterns in neuroscience. 

\textbf{Protein engineering.}
Protein engineering is at the forefront of biotechnology, presenting a transformative potential in diverse domains such as antibody design, drug discovery, food security, ecology and beyond. Taking advantage of extensive protein databases, machine learning models have significantly accelerated the pace of protein engineering and directed evolution. Recently, TDA approaches have been used to dramatically reduce the structural complexity of mutant and wide-type proteins~\citep{qiu2023persistent}. TDL based on persistent homology has shown to be a top performer in an extensive study involving 34 datasets, but a persistent Laplacian approach outperforms persistent homology in protein engineering applications~\citep{qiu2023persistent}. TDL has also facilitated deep mutational scanning~\citep{chen2023topological}.

\textbf{Chip design.} Chip design has been accelerated by the combination of machine learning with geometric and topological methods. In this field, the central element is the so-called netlist, which is a network comprising cells (logic gates) and nets (connections among cells). Netlists can be effectively modeled as (directed) hypergraphs, where nodes and hyperedges represent cells and nets, respectively. A key goal in chip design is to efficiently lay out netlists within a specified 2D area. The aim is to optimize several properties, such as minimizing total wire length and reducing congested areas known as `hotspots'. However, optimizing netlists, including their placement and routing, is a complex and time-consuming task. It involves multiple steps (including placement, global routing, detailed routing) and repeated iterations. This complexity requires data-driven methods that can speed up these processes~\citep{Kahng23,Mirhoseini2021}. For example, there is growing interest in predicting the properties of a synthesized netlist without undergoing the lengthy placement and routing process~\citep{Ghose2021,Xie2021,YY22}. However, netlists pose several challenges for existing graph learning models. Such models are often large, containing millions of nodes, and the long-range interactions and path information are crucial for determining netlist properties, such as timing and routability. Moreover, various symmetry structures in netlists must be considered. These complexities highlight the need to integrate topological and geometric concepts into the learning process.~\citet{Luo2024} have made a step towards this direction by integrating persistent homology and spectral features within a carefully designed neural network based on directed hypergraphs to predict netlist properties. 

\textbf{Semantic communications.} A key challenge in communication systems is balancing the complexity of data representation with the significance of symbols transmitted to convey the intended meaning or semantics within an allowable margin of error or distortion. Efficiently representing semantic knowledge involves mapping out the relations between the elements of a language, broadly defined, by creating a corresponding topological space~\citep{barbarossa2023}. TDL can play a crucial role in semantic communications, particularly in extracting relational semantic features from data. This is vital for transmitting information that is strictly necessary to convey the desired meaning or to execute a specific task, such as inference, control, or actuation, while meeting certain performance criteria, such as latency, energy efficiency, and accuracy. The integration of TDL in this context has the potential to revolutionize the way of understanding and optimizing the transmission of meaning in communication systems.

\textbf{Satellite imagery.} Satellite imagery, typically characterized by noise, sparsity, and a plethora of geophysical features on different spatial scales, presents unique challenges for data analysis.  Currently, CNNs are the preferred choice for creating automated detection models that identify phenomena in satellite imagery. However, a CNN designed for a specific spatial scale might not be effective for images of different scales. While there are several methods that address multiscale processing, such as hierarchical CNNs~\citep{he2022} and vision transformers~\citep{bazi2021}, TDL offers an alternative approach to multiscale modeling. The use of topological invariants in this context can be particularly beneficial for several reasons. For example, topological invariants are not affected by geometric transformations such as translation, rotation, scaling, or shearing, making them highly robust and reliable for satellite imagery analysis~\citep{Ver_Hoef2023}. This attribute of TDL can potentially enhance the accuracy and efficiency of analyzing complex satellite data.~\citet{DWW19} have used topological techniques to extract and analyze features from satellite images, showcasing the practical effectiveness of TDL in real-world scenarios.

\textbf{Materials science.}
Using topological properties in experimental materials data analysis presents a useful approach to streamline the complex processing-structure-property relation in materials, as highlighted by~\citet{NA2022,Papamarkou2022}. Embedding material structure into topology-based supervised or unsupervised learning algorithms reveals crucial features in intricate processing-structure-property relations, for example, in datasets such as those found in high-entropy alloys~\citep{SPANNAUS2021}. Taking advantage of topological crystallographic properties~\citep{sunada2013}, atom-specific machine learning models based on persistent homology have been developed, demonstrating improved accuracy in predicting the formation energy of crystalline compounds~\citep{jiang2021topological}. TDL may have the potential to uncover latent properties in quantum or other materials because it takes into account the underlying topological properties, such as symmetry, in the structure of the materials.

\end{document}